# Privacy-preserving household load forecasting based on non-intrusive load monitoring: A federated deep learning approach

Xinxin Zhou [1], Jingru Feng [1,*], Jian Wang [1], Jianhong Pan [2]

[1] School of Computer Science, Northeast Electric Power University, Jilin 132012, China

[2] State Grid Jilin Electric Power Company Limited, Changchun13000 , China

* Corresponding author. E-mail address: 1665639965@qq.com.

**ABSTRACT:** Load forecasting is very essential in the analysis and grid planning of power systems. For this reason, we first propose a household load forecasting method based on federated deep learning and non-intrusive load monitoring (NILM). For all we know, this is the first research on federated learning (FL) in household load forecasting based on NILM. In this method, the integrated power is decomposed into individual device power by non-intrusive load monitoring, and the power of individual appliances is predicted separately using a federated deep learning model. Finally, the predicted power values of individual appliances are aggregated to form the total power prediction. Specifically, by separately predicting the electrical equipment to obtain the predicted power, it avoids the error caused by the strong time dependence in the power signal of a single device. And in the federated deep learning prediction model, the household owners with the power data share the parameters of the local model instead of the local power data, guaranteeing the privacy of the household user data. The case results demonstrate that the proposed approach provides a better prediction effect than the traditional methodology that directly predicts the aggregated signal as a whole. In addition, experiments in various federated learning environments are designed and implemented to validate the validity of this methodology.

## 1. Introduction

Since AC power cannot be stored, the power generation capacity of the power plant must reach a stable dynamic balance with the power consumption [1], in order to guarantee that the power grid can operate stably and efficiently, reasonable power supply plans and power dispatching methods need to be developed, and accurate power forecasting is a necessary condition. With the continuous growth of the world's population, the demand for energy from mankind is increasing day by day. At present, domestic dwellings account for about 20-40% of the world's total energy use, and the household demand load often has a significant impact on the peak load of seasonal and daily electricity [2]. Due to the inherent uncertainty of distributed renewable energy resources [3, 4], the renewable integration also brings new challenges to household load forecasting. Accurate forecasting of household electricity demand is not only a prerequisite for guaranteeing the safety and stability of the power grid, but also for peaking and flattening valleys by means of floating electricity prices and balancing the supply and demand relationship at peak electricity consumption.

### 1.1. Literature review

At present, load forecasting methods commonly used abroad can be separated into statistical methods (e.g., the non-linear extrapolation method [5-7], time series method [8-10], etc.) and machine learning forecasting methods (e.g., artificial neural networks (ANN) [11-14], support vector machines (SVM) [15-17], etc.). Since a single forecasting method often has various defects, the combined forecasting method is gradually applied to the research of power system load forecasting, and its performance is better than all independent load forecasting models [18]. Reference [19] proposed a load forecasting method based on user clustering. After clustering historical load curves, wavelet analysis is proposed to predict the future 24-hour load curve. Reference [20] compared the performance of 7 existing load forecasting technologies, including linear regression, ANN, SVM, and their variants on two commercial buildings and three residential household data sets. The results show that these traditional system-level load forecasting methods can provide good forecasts for commercial buildings, but perform poorly on household data. As for load prediction at the household level, there are still few reference works, and most of the existing forecasting methods directly predict the total power, such as Kalman filter [21-23], Autoregressive comprehensive moving average model [20][24, 25], etc.

Research on household or residential electricity demand estimation currently takes several different ideas, including evaluating and improving existing load forecasting techniques, developing and researching new methods, or a combination of the two. Reference [26] introduced a deep recursive neural network to learn and correlate shared information between customers and solve the problem of overfitting. The performance of this method on the Irish housing data set exceeds the autoregressive integrated moving average model (ARIMA), SVM, and classic recurrent neural networks. Reference [27] proposed a convolutional-long short-term memory neural network (CNN-LSTM) model with selective autoregressive characteristics to enhance single-step-ahead power load forecasting. This improvement can be observed in the three spatial granularities of apartments, floors, and building floors and achieves higher prediction accuracy. Recently, some scholars have proposed a household prediction method based on NILM, which can be more accurately used for household-level power prediction. The reason is that certain equipment used alone may have a strong time dependence, such as air conditioning. In summer or winter, we turn on the air conditioner but it is not turned on 24 hours a day. In spring or autumn, it is only turned on occasionally, which means that the air conditioner changes over time. When different appliances are aggregated, the time dependence of this single appliance may be broken. Therefore, predicting the total power distribution of the entire family may be less accurate than predicting the power distribution of a single appliance. Reference [28] proposed a method for household electricity consumption prediction based on NILM, which combines the correlation of electrical behavior in its state duration, and the experiment has good accuracy, but this method does not consider time comprehensively information. Reference [29] developed a NILM algorithm for analyzing the operating characteristics of different appliances. Reference [30] proposed a new method for single-household electricity consumption prediction based on NILM and aggregated spectral clustering, and used four public data sets for testing. However, this method has a high error, does not take advantage of deep learning, and does not describe the part of data preprocessing. Such deep learning based prediction methods have also been the subject of research in recent years since rich training data yield the expected model performance and deep learning based methods have a great number of trainable parameters. At present, there are very few existing researches on this kind of household-level load forecasting based on NILM.

In addition, if the predictive model is to be applied to a wider range of electricity consumers and appliances, sufficient electricity demand data must be collected from multiple data owners, but multi-source load data may cause privacy disputes [31]. In practice, some households may be reluctant to share data with others, especially when power data information is stored in a central server or during transmission. Due to the frequent exchange of information, it is easy for a third party to obtain power data [32]. Currently, many countries have promulgated specific laws. For instance, the "Network Security Law" promulgated in 2017 clearly requires that the collected user information should be strictly confidential [33]. Thus, to resolve the above-mentioned questions and achieve a balance between consumer privacy protection and data utilization, federated learning is proposed. Federated learning does not need to collect raw data, that is, the private data of family users does not need to be uploaded to the central server for training. Each family participant can conduct model training locally, and finally, just upload the updated model parameters to the cloud server [34]. The framework proposed by federated learning balances the relationship between data utilization and privacy protection, and in the process improves the performance of the co-training model. At present, there is also a very small part of the literature discussing the application of federated learning in electric energy. For example, reference [31] proposed a federated learning method for power consumer feature identification. This method was composed of privacy protection principal component analysis and FL-ANN model. This model has been proven to perform better. The reference [35] used FL-LSTM for energy demand forecasting, preserving the data privacy of energy consumers, and compared a centrally trained approach with a localized training approach. In addition, how to apply federated learning to the literature at the household load level is also an urgent and challenging issue.

### 1.2. Contribution of this paper

In view of the above, most of the current household load forecasts are directly based on the total power forecast as a whole, which will cause large errors. As for the few integrated prediction methods based on NILM, there are also a series of problems such as low prediction accuracy, lack of data preprocessing, not taking advantage of deep learning, or not considering the comprehensive connection of time. In addition, the centralized model is vulnerable to external attacks, which may lead to third parties having the opportunity to obtain load data. In particular, load information may be destroyed when stored in a centralized server or during transmission.

This paper presents for the first time a novel federated deep learning prediction model based on NILM. This method integrates federated learning and Bi-directional long short-term memory network-Attention mechanism (BiLSTM-Attention). The following are the main contributions:

(1) This paper presents a load prediction method based on federated deep learning (FedDL), which combines federated learning and BiLSTM-Attention. For all we know, this is the first study of federated learning in household load forecasting based on NILM. In this prediction method, under the precondition of guaranteeing the accuracy of the forecast, the data of the users in the house is not leaked, and the data can be isolated to ensure that the data of the users in the house is not leaked, and meet the needs of user privacy protection and data security. On the precondition of guaranteeing the accuracy of the prediction, while reducing the data security risk, it can also learn from the electrical power of other households.

(2) A method based on NILM to predict user household energy consumption is proposed. The NILM technology extracts a single load pattern from the historical total load demand, avoiding the strong time dependence of a single device. Experiments demonstrate that the integrated prediction method based on NILM has better prediction results than the traditional method that directly predicts the aggregated signal as a whole.

(3) Use the data decomposed by the CNN-LSTM hybrid deep learning model to test the proposed federated deep learning prediction model to validate the validity of the proposed model; In addition, experiments in various federated learning environments are designed and implemented to validate the validity of this methodology.

### 1.3. Organization of this paper

The main points of this paper are as follows: Section 1 introduces the relevant background and relevant work on household load forecasting; Section 2 presents the associated theories of machine learning; Section 3 mainly proposes the federated deep learning model, and introduces the specific process of training this model. Section 4 performs the relevant training and testing of the proposed FedDL model using the data obtained based on the NILM, and analyzes the test results. Section 5 summarizes this article.

## 2. Related preliminary work

### 2.1. Convolutional neural network (CNN)

Due to their powerful learning and generalization capabilities, artificial neural networks are widely used to deal with pattern recognition problems in the engineering field [36, 37]. This article uses CNN to extract data features in the data preprocessing part. Eq. (1) is the output of the convolutional layer:

$$a_j^l = \mathrm{conv}\left(\sum w_{ij}^l X_i^{l-1} + w_b\right) \qquad (1)$$

where $w_{ij}^l$ represents the weight of the convolutional layer; $w_b$ indicates the bias term of the filter; $a_j^l$ and $X_i^{l-1}$ are the input and output in the convolution process [38], separately. The nonlinear activation function is:

$$X_j^l = f\left(a_j^l\right) \qquad (2)$$

$$X_j^{l+1} = \mathrm{pooling}\left(X_j^l\right) \qquad (3)$$

The pooling function represents the pooling operation, $X_j^{l+1}$ represents the output after the pooling layer.

## 2.2. Bi-directional long short-term memory network-attention mechanism

### 2.2.1 Long short-term memory network (LSTM)

LSTM is a typical recurrent neural network, which is particularly suitable for dealing with time-series related pattern recognition problems. Fig. 1 shows the LSTM structure.

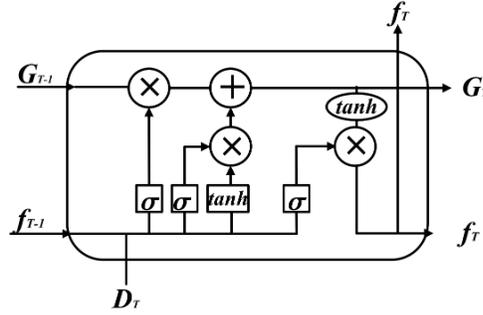

Fig. 1. LSTM structure diagram

The input parameters are the memory cell state $G_{T-1}$ and hidden layer state $f_{T-1}$ at time $(T-1)$, and the input sequence element value $D_T$ at time $T$. The output parameters are the memory cell state $G_T$ and the hidden layer state $f_T$ at time $T$. The corresponding calculation equation in Eq. (4-8):

$$r_T = \sigma\left(W_i\left[f_{T-1}, D_T\right] + h_i\right) \quad (4)$$

$$l_T = \sigma\left(W_f\left[f_{T-1}, D_T\right] + h_f\right) \quad (5)$$

$$g_T = \tanh\left(W_c\left[f_{T-1}, D_T\right] + h_c\right) \quad (6)$$

$$G_T = r_T * g_T + l_T * G_{T-1} \quad (7)$$

$$P_T = \sigma\left(W_o\left[f_{T-1}, D_T\right] + h_o\right) \quad (8)$$

$$f_T = P_T * \tanh(G_T) \quad (9)$$

where $W_i$, $W_f$, $W_o$, $W_c$ represent the weight matrix of input gate, forget gate, output gate, and memory unit state respectively; $h_i$, $h_f$, $h_o$, $h_c$ represent the bias terms of each gating unit and memory unit [39]; $\sigma$ represents the sigmoid function [40]; * means that the corresponding elements of the two vectors involved in the operation are multiplied.

### 2.2.2 BiLSTM

The BiLSTM network has the function of capturing the characteristics of the information before and after [41]. Since in the actual power system, the power load data is a time-varying series of data and non-linear, its load at a certain moment is not only influenced by other factors (such as holidays and social environment), but also simultaneously associated with the past input features, while the future input features can also reflect the present load features to some extent [42].

Therefore, in order to effectively obtain the time-series variation features about time, BiLSTM is used to mine the intrinsic connection between the current data and the data of past and future moments, which is shown in Fig. 2.

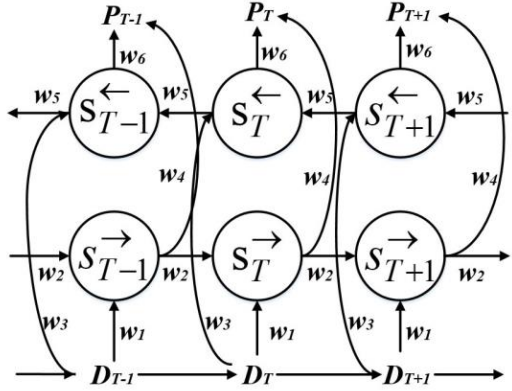

Fig. 2. BiLSTM structure diagram

The BiLSTM hidden layer state $P_T$ at moment $T$ can be found by the forward hidden layer state $\overrightarrow{s_T}$ and the backward hidden layer state $\overleftarrow{s_T}$ in two parts. The forward hidden layer state $\overrightarrow{s_T}$ is decided by the current input $D_T$ and the forward hidden layer state $\overrightarrow{s_{T-1}}$ at moment $(T+1)$. The backward hidden layer state $\overleftarrow{s_T}$ is decided by the current input and the backward hidden layer state $\overleftarrow{s_{T+1}}$ at moment $(T+1)$. The calculation formula is shown below, where $w_i$ ($i=1, 2..., 6$) is the weight from one cell layer to another cell layer.

$$\overrightarrow{s_T} = f\left(w_1 D_t + w_2 \overrightarrow{s_{T-1}}\right) \tag{10}$$

$$\overleftarrow{s_T} = f\left(w_3 D_T + w_5 \overleftarrow{s_{T+1}}\right) \tag{11}$$

$$P_T = g\left(w_4 \overrightarrow{s_T} + w_6 \overleftarrow{s_T}\right) \tag{12}$$

### 2.2.3 Attention mechanism

The attention mechanism highlights key information through weight distribution to ensure that features are not lost in the model training process, so it can more effectively mine the long-distance data features of time series with correlation [43]. The attention mechanism takes the timing sequence of the input power load-related data as the reference object of the neural network [44]. Combined with BiLSTM, the neural network spontaneously selects to retrieve the information from the BiLSTM. Its basic structure is shown in Fig. 3.

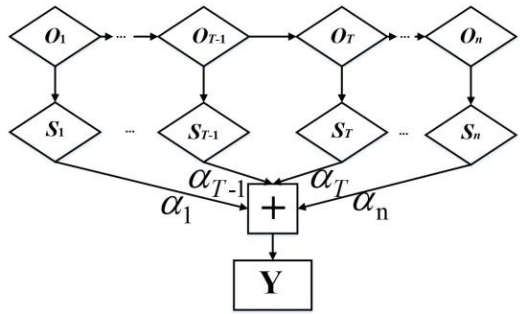

Fig. 3. Structure diagram of attention mechanism

$O_T$ is the $T$-th feature vector of BiLSTM network output, which is input into the hidden layer of attention mechanism to

obtain the initial state vector $S_T$, and then the final output state vector $Y$ is obtained by multiplying and summating the corresponding weight coefficient $\alpha_T$. The calculation formula is as follows:

$$e_T = \tanh(w_T S_T + b_T) \tag{13}$$

$$\alpha_T = \frac{exp(e_T)}{\sum_{i=1}^{T} e_i} \tag{14}$$

$$Y = \sum_{T=1}^{n} \alpha_T S_T \tag{15}$$

where $w_T$ represents the weight coefficient matrix of the $T$-th feature vector, and $b_T$ denotes the offset corresponding to the $T$-th feature vector [45].

## 2.3 Federated learning

Federated learning is a distributed training method that uses data sets scattered among the participants, integrates data information from multiple parties through privacy protection technology, and collaboratively builds a global model [46]. During the model training process, the relevant information about the model (such as model parameters, model structure, parameter gradient, etc.) can be exchanged among participants (The exchange mode can be plaintext, data encryption, adding noise, etc.) [47], but the local training data will not leave the local. This exchange does not expose local user data, decreasing the danger of data leakage. Trained federated learning models can be shared and deployed across data participants.

The training process can be organized into three steps:

(1) Task initialization: the server determines the training target, parameters, and participating terminal devices, and passes the global model $\omega_G^0$ to the participating equipment.

(2) Local model training and update: global model $\omega_G^t$ downloaded from the server, where $t$ represents the current training round. The participant appliance $i$ trains the model locally and updates the parameter $\omega_i^t$ this time. The updated local model parameter is expressed as $\omega_i^a$, the goal of training is to find the best local model by minimizing the loss function $L(\omega_i^t)$ and upload it to the server:

$$\omega_i^a = arg_{\omega_i^t} min\left[L(\omega_i^t)\right] \tag{16}$$

(3) Global model aggregation and update: The server receives the model parameters uploaded by the participant devices and aggregates them, and finally updates the global model $\omega_G^{t+1}$:

$$L(\omega_i^t) = \frac{1}{N} \sum_{i=1}^{N} L(\omega_i^t) \tag{17}$$

The above training process is iterated until the convergence or termination condition training technique is satisfied. From the classical training model of federated learning, we can find that the local appliance communicates with the server through model parameters during training, so as to protect privacy.

## 3. Proposed prediction model

In this part, the process of each part is explained in detail. Firstly, the data is preprocessed, and the individual load of household appliances is decomposed from the total load demand through the NILM method based on CNN-LSTM hybrid model. Then the power of individual appliances is predicted separately by the FedDL prediction model, and the predicted power of individual appliances is aggregated to form the overall power prediction. Finally, the validity of the proposed model is evaluated by some evaluation indicators.

### 3.1 Data decomposition

In this paper, the NILM method is based on CNN-LSTM hybrid model to decompose the total load data of the training sample and obtain the energy consumption value of individual appliances. Fig. 4 presents the structure of the CNN-LSTM.

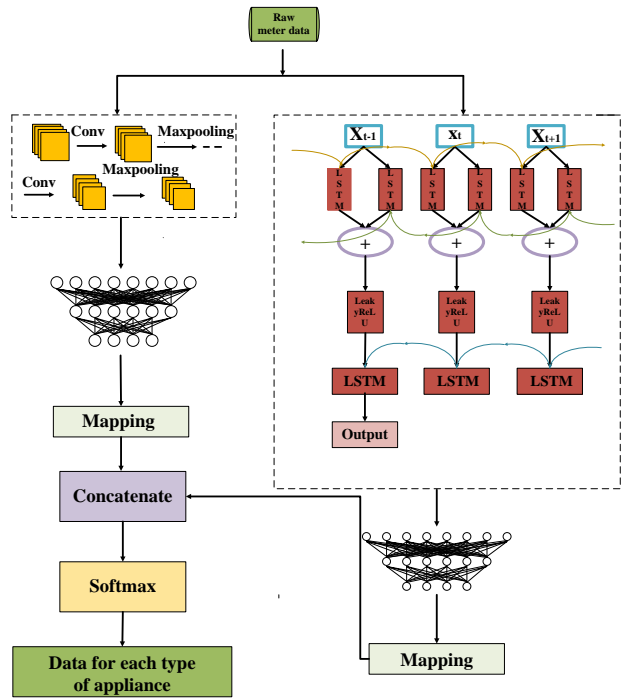

Fig. 4. Structure diagram of CNN-LSTM

Firstly, the CNN and LSTM are trained separately using the pre-processed data, and then the feature information extracted from each of the two networks is handled as the equal dimension using the mapping layer, and finally, the two are concatenated. For more specifics, please refer to reference [48].

### 3.2 FedDL-based prediction model

In order to build a better neural network, a large amount of data containing various devices and responding to consumer behavior habits needs to be fed to the network. However, during model training, the local data owner may encounter the risk of privacy leakage and loss of control over the data. To address this potential risk, we use federated learning, which is in sharp contrast to traditional centralized machine learning techniques where all sample data is uploaded to a single server. The flow chart of the proposed FedDL prediction model is shown in Fig. 5.

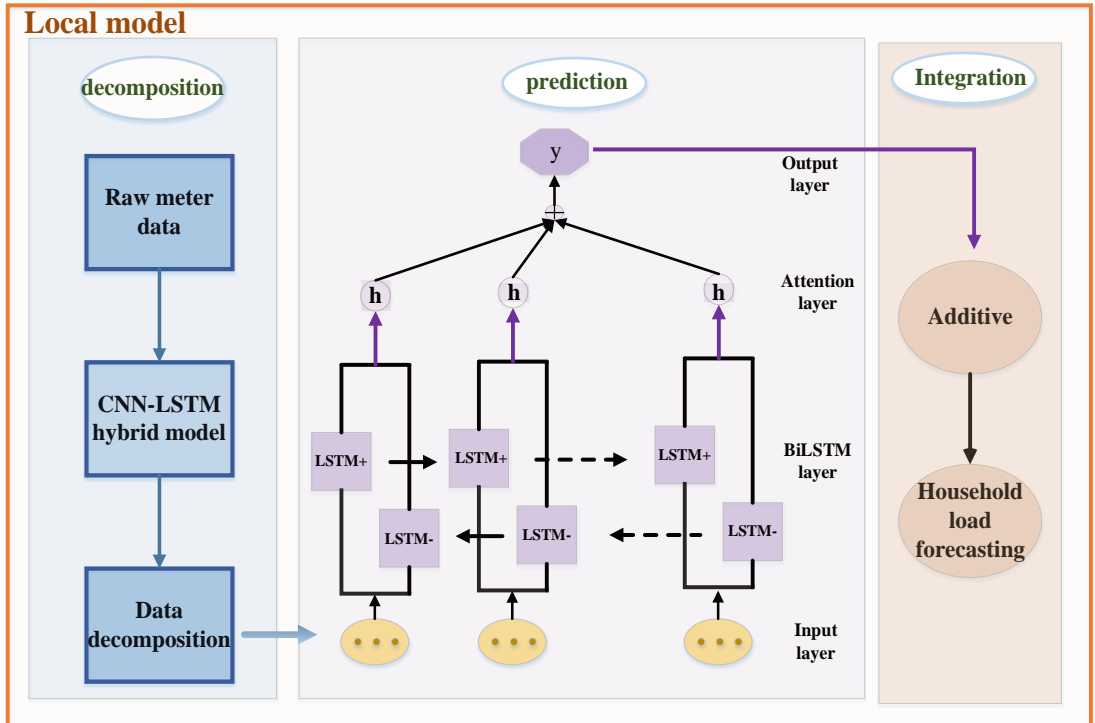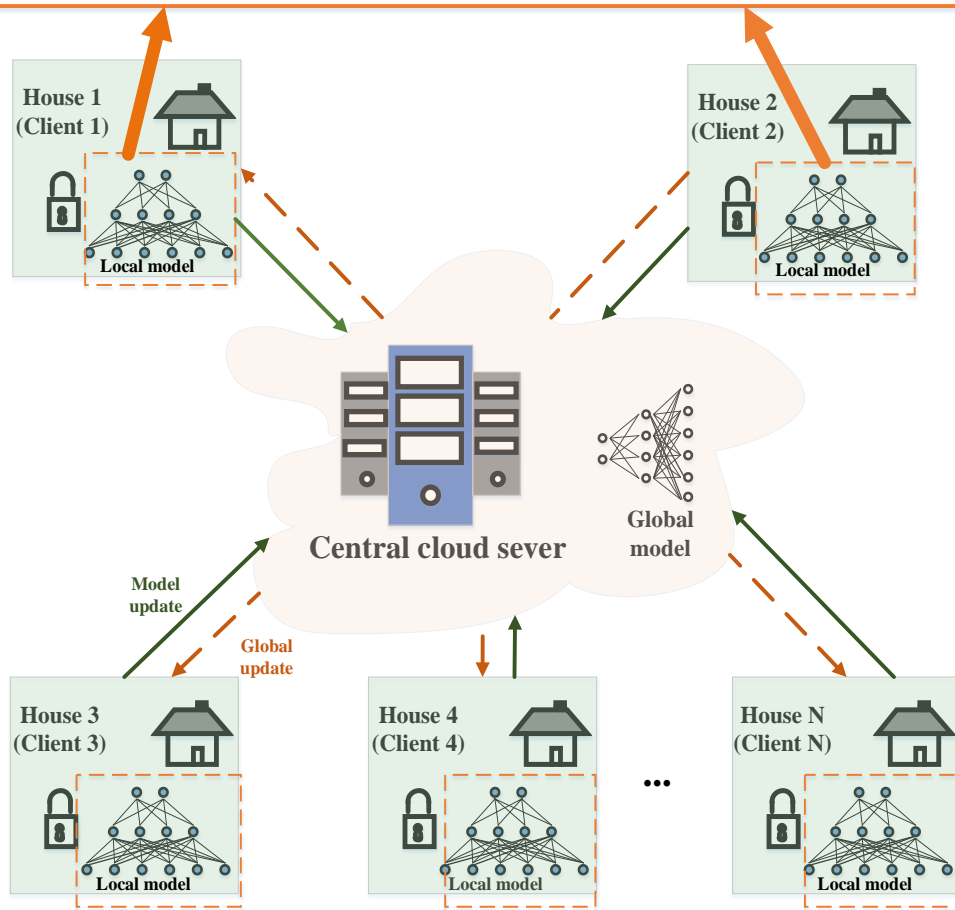

Fig. 5. Flowchart of the FedDL prediction model

For the purpose of illustrating, we assume that each household has the same computational power, the same hyperparameters (e.g., learning rate, minibatch size) for training in FedDL, and the same optimization algorithm is used. As shown in Fig. 5, firstly, the initial global model will deploy its parameters to the client. After receiving the parameters of the global model, the local household will train its local deep learning model based on local data and update its parameters, and then transmit these parameters to the global server. The global deep learning model receives the locally updated parameters and adopts a federated average for the parameters, and then updates the global deep learning model with the results based on the federated average. After that, the updated global parameters will be passed to the local model of each house, and the final global deep learning model will be generated after all the global communication rounds have been run between the local residents and the central server.

We explained it step by step under the local model. The individual load of household appliances is decomposed from the total load demand through the NILM method based on CNN-LSTM hybrid model. Then the power of individual appliances is predicted separately by the prediction model, and the predicted power of individual appliances is aggregated to form the overall power prediction.

### 3.3 Local model structure parameters

The specific internal structure is shown in Fig. 6. First, the model delivers the pre-processed sequence data to the forward and the backward LSTM hidden layer, and the two are combined with the output vector as the bi-directional timing feature vector of the electric load at the moment *T*. The attention layer is responsible for assigning attention weights, highlighting the key factors, ignoring irrelevant information, and continuously iterating the optimal weight parameter matrix. Then, in order to obtain the new vector of inputs to the fully connected layer, the weight vector is combined with the shallow layer output. Finally, the predicted load values are obtained by the fully connected layer (sigmoid as the activation function).

The LSTM layer can remember or forget significant information. In principle, the more layers, the stronger the learning ability. But too many layers will lead to difficult convergence and a long training time. Therefore, in the model of this paper, a two-layer LSTM is set. The number of neurons in the first layer is set as 128, and the number of neurons in the second layer is 68.

### 3.4 Evaluation indicators of federated deep learning model

In machine learning, there are two major problems: classification and regression. The essence of both is to predict the input. The difference between the two lies in the type of output results. The output of the classification problem is a qualitative discrete variable, while the output of the regression problem is a quantitative continuous variable. For these two kinds of problems, there are corresponding evaluation indexes to measure the validity of the models and algorithms. As a typical regression problem, power load prediction mainly has the following evaluation indexes:

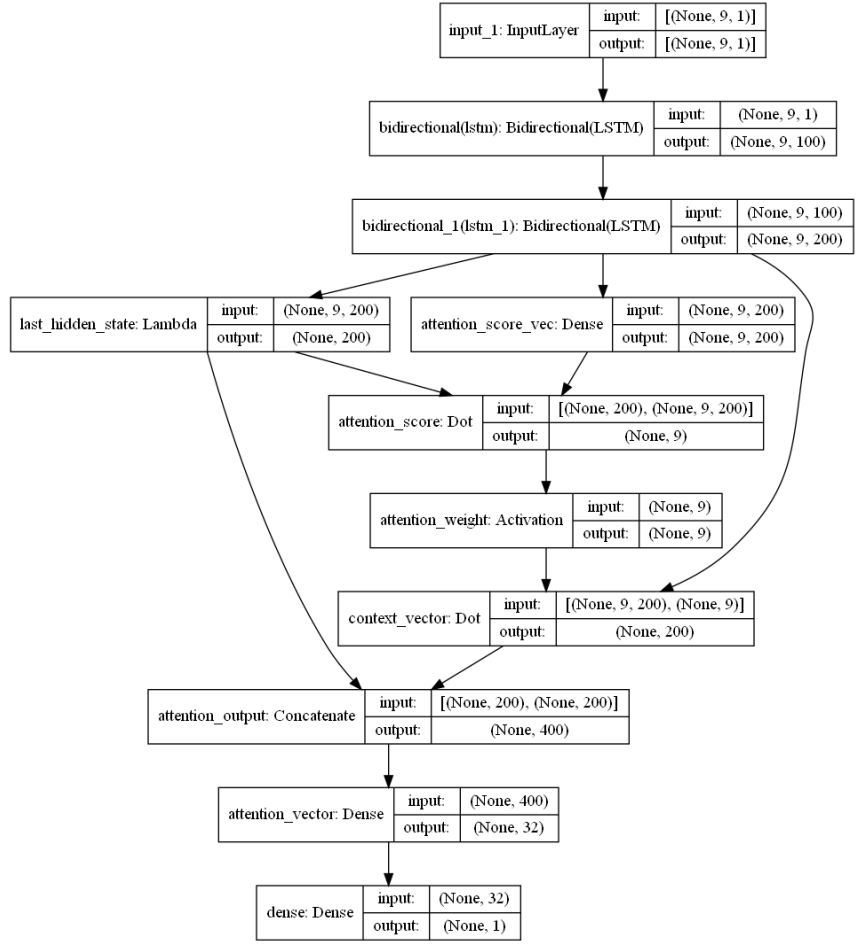
Fig. 6. BiLSTM-Attention internal structure diagram

(1) Mean Absolute Error (MAE): The advantage is that the calculation is simple and the description error is clear. The specific formula is shown in Eq. (18):

$$\text{MAE} = \frac{1}{n}\sum_{i=1}^{n}\left|y_{tru} - y_{pre}\right| \quad (18)$$

(2) Root Mean Square Error (RMSE): For data whose unit measurement is too large, this operation can retain the original description unit for the data. For example, if the unit of electricity load is KW, the square of the difference will increase by an order of magnitude, which will add trouble to the model description, so RMSE is chosen to maintain the original magnitude. The smaller the original root mean square error, the more accurate the prediction. The specific formula is shown in Eq. (19):

$$\text{RMSE} = \left(\frac{1}{n}\sum_{i=1}^{n}(y_{tru} - y_{pre})^2\right)^{\frac{1}{2}} \quad (19)$$

In Eqs. (18) and (19), $y_{tru}$ and $y_{pre}$ are the true and predicted values, respectively.

## 4. Case study

In this part, the integrated power is decomposed into individual device power by the NILM method based on CNN-LSTM hybrid model, and the power of each device is predicted separately using the federated deep learning model. Finally, the predicted power values of individual appliances are aggregated to form the overall power prediction. The validity of the method is confirmed through qualitative and quantitative evaluation results. In addition, experiments in various federated learning environments are designed and implemented to validate the validity of this methodology. The programming language in the experiment is python and Tensorflow2.0 is used.

In the BiLSTM-Attention model used in this paper, a 2-layer LSTM is set. The number of neurons in the first layer is set as 128, and the number of neurons in the second layer is 68. The activation function is tanh, the learning rate is 0.001, Adam is selected as the optimization function, and the batchsize is 512.

### 4.1 Generation of the data set

The data set UKDALE established by British scholars Jack Kelly and William Knottenbelt is used for the experiment [42]. The load data set contains information about individual loads in five households as well as the total household electricity consumption. And use the load data from July 1, 2013 to July 3, 2013 of five households in the UKDALE dataset as the data set used in this study. The previous 4070 minutes of data were used as the training set, and the remaining 250-minute data is employed as the testing set. The data in the dataset is sampled every 6 seconds.

### 4.2 Experimental results

For ease of analysis without loss of generality, the experimental results shown in this article take household 1 as an example. The experimental results are shown below.

Using CNN-LSTM hybrid model decomposition method to decompose the total load data of the training samples to obtain the energy consumption values of each load device, without considering the simultaneous existence of multiple electrical appliances. The load decomposition results of loading equipment for a certain three days in April are shown in Fig. 7. This step of decomposition sets the stage for the next step of prediction.

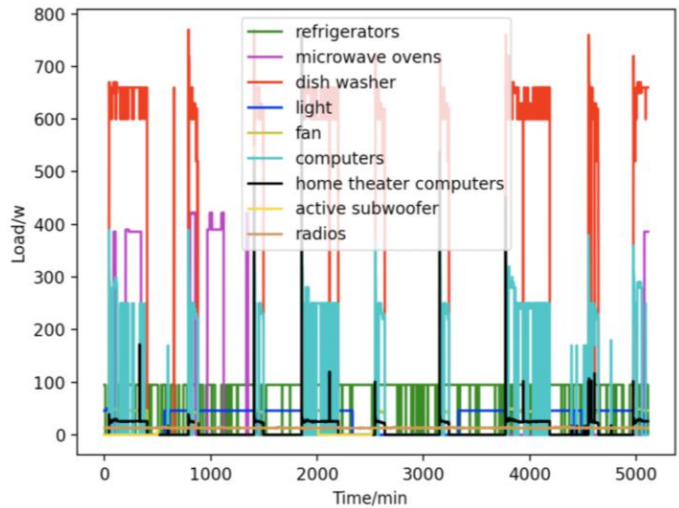

Fig. 7. Load curve

In Fig. 7, the 9 kinds of electrical appliances obtained by decomposition are refrigerators, kettles, home theater computers, microwave ovens, radios, fans, light, active subwoofers, and computers. The power curves of the 9 electrical appliances are marked by different colors.

Based on the load decomposition results of nine load devices, nine load prediction models are trained respectively according to the prediction model structure proposed in Section 3. Figs. 8-16 show the real values of 9 appliances compared to the predicted values, and Fig. 17 shows the total predicted value of the household loads.

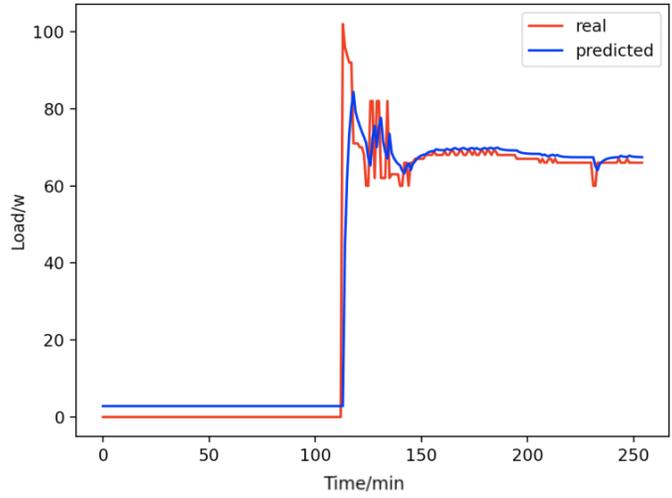

Fig. 8. Active subwoofer

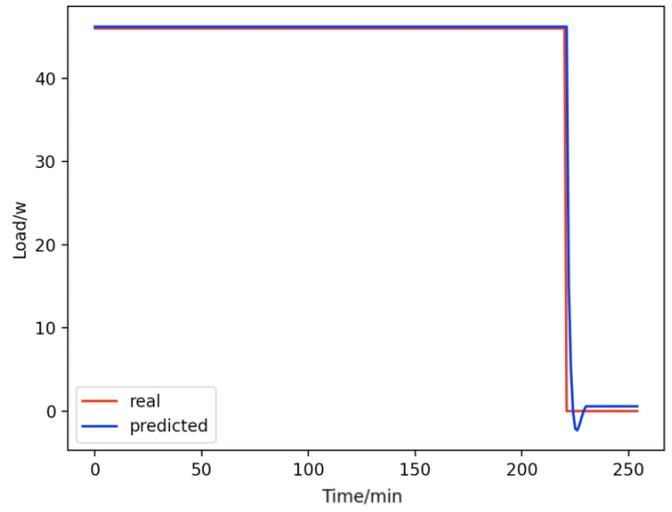

Fig. 9. Light

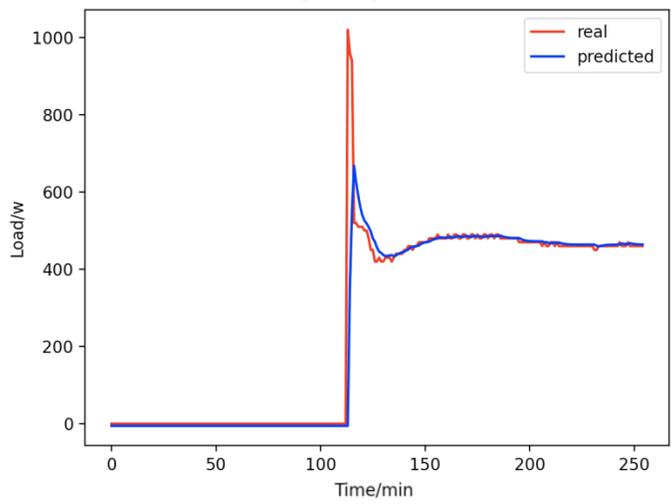

Fig. 10. Home theater computers

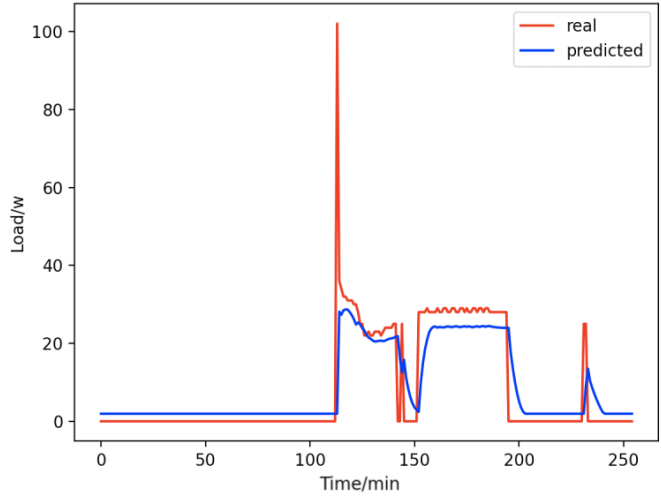
Fig. 11. Fans

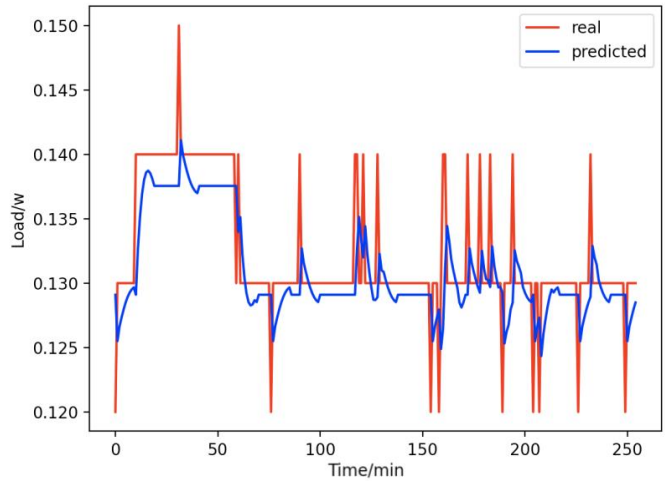
Fig. 12. Radios

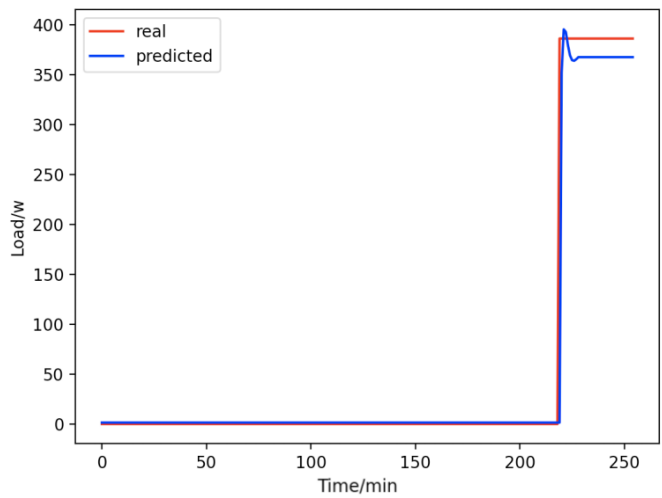
Fig. 13. Microwave ovens

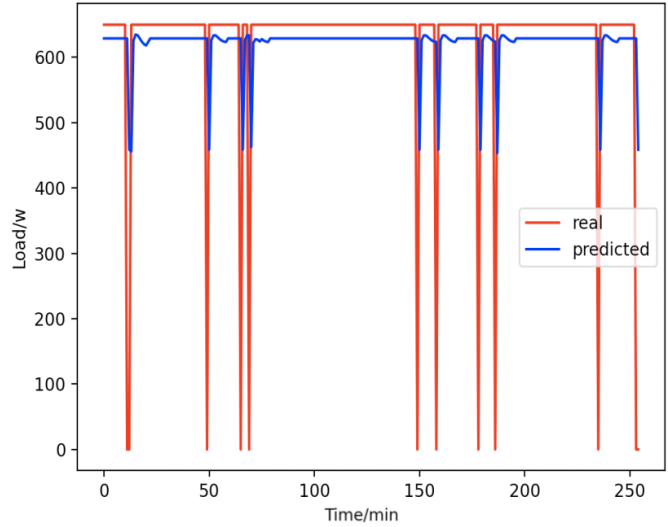

Fig. 14. Kettles

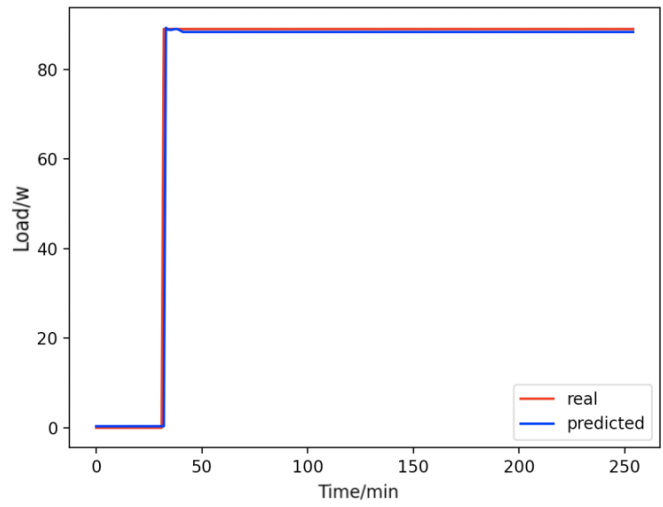

Fig. 15. Refrigerators

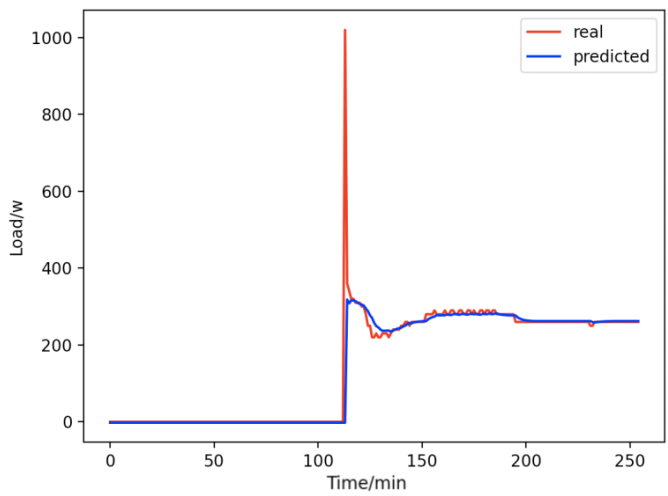

Fig. 16. Computers

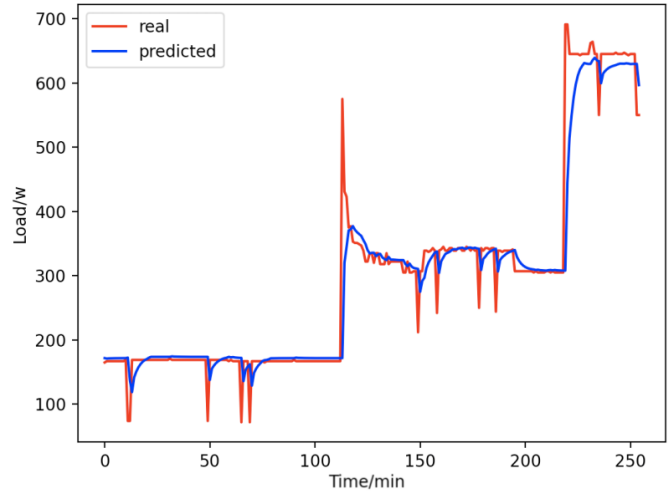

Fig. 17. Total forecast

From Figs. 8-16, we can see that the results of the true and predicted values of each appliance are very close, and Fig. 17 illustrates the total household load forecasting results. Its total true value is also very close to the predicted power with better results. These results demonstrate that the NILM-based load decomposition approach is effective and the FedDL model manages to accurately forecast the household loads.

To prove that the NILM-based integrated prediction methodology provides a better prediction effect than the traditional methodology that directly predicts the aggregated signal as a whole, we conducted a comparative experiment based on integrated prediction after load decomposition and overall direct prediction.

Fig. 18 shows the comparison between the integrated prediction based on load decomposition and the overall direct prediction. It is observed that both can reflect the electricity consumption trend of household users, but the method based on integrated prediction and comprehensive prediction has a better fitting ability to the actual load curve. In contrast to the prediction results based on the overall direct prediction, in the overall equipment, different equipment has different power characteristics. In addition, there may be strong time dependence in the power signal of single equipment, and the overall direct prediction ignores the difference in power characteristics between devices, resulting in a large mistake in the overall direct prediction method. However, the load data based on load decomposition can avoid the above problems.

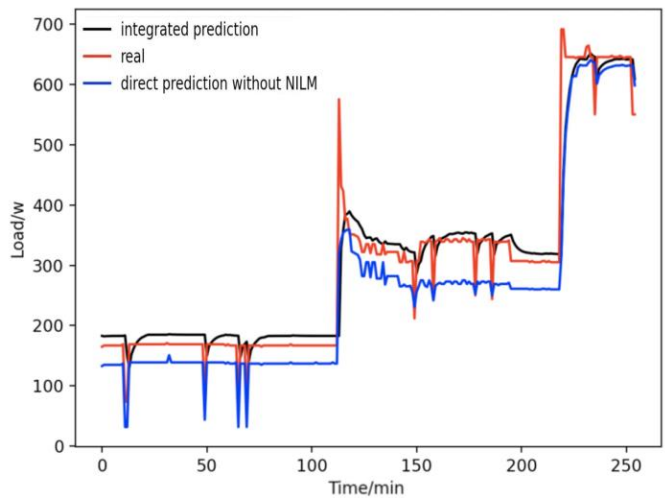

Fig.18. Comparison of overall direct forecast and individual forecast

## 4.3 Comparisons with other methods

To reasonably evaluate the performances of the proposed approach, comparisons with other methods have been carried out. For the different models selected in this paper, related experiments are done on the basis of NILM load decomposition. The same input samples are used for training the model, and the RMSE and MAE are used as the evaluation indexes of the prediction model.

As can be seen from Table 1, the performance of the FedDL is better than that of all other alternative algorithms (besides the BiLSTM-Attention). However, the FedDL provides better privacy protection for co-modeling participants than the BiLSTM-Attention. In this prediction method, under the precondition of guaranteeing the accuracy of the prediction, the data of household users are not leaked.

Table 1 Comparison of evaluation indexes

|  | MAE | RMSE |
| --- | --- | --- |
| Proposed method | 0.08141 | 0.16739 |
| BiLSTM-Attention with NILM | 0.07825 | 0.15956 |
| ANN with NILM | 0.28376 | 0.34675 |
| LSTM with NILM | 0.10956 | 0.18266 |
| FFANN with NILM | 0.27869 | 0.50923 |
| ARIMA with NILM | 0.28865 | 0.55243 |
| SVM with NILM | 0.28914 | 0.52826 |

To further validate the validity of the proposed model in a more profound way, we also test it on the other 4 families, where the same amount of appliances and data are selected as household 1 (All the selected models are based on NILM load decomposition). Meanwhile, the first three models in Table 1 are selected for comparative experiments, which are shown in the figure below.

Figs. 19-21 provide the comparative error analysis of the five households in the UKDALE data set under different models. As for the comparison with the BiLSTM-Attention, the performance of the proposed model is slightly inferior to that of the BiLSTM-Attention, but the FedDL can effectively protect user privacy. In terms of these two indicators, the FedDL outperforms the LSTM and ANN, which further proves the superiority of the proposed approach.

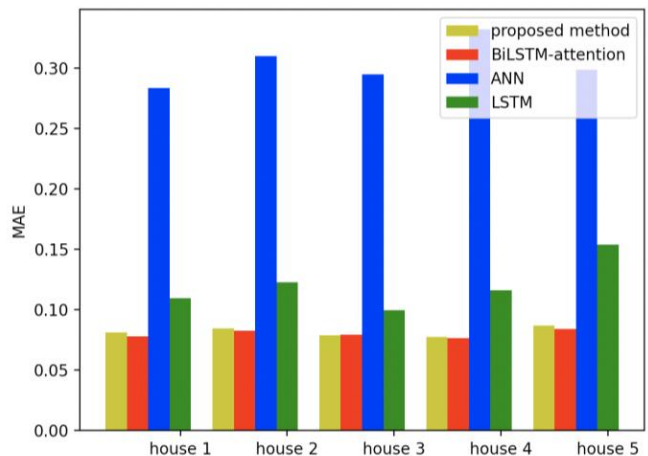

Fig. 19. Comparison of 5 households using MAE using different methods

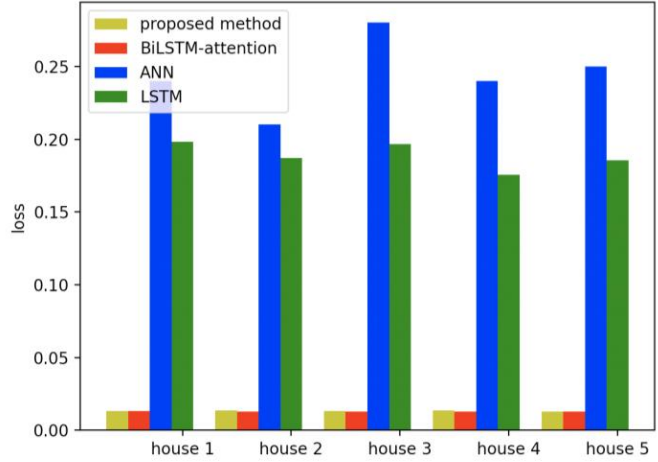

Fig. 20. Comparison of 5 households using loss using different methods

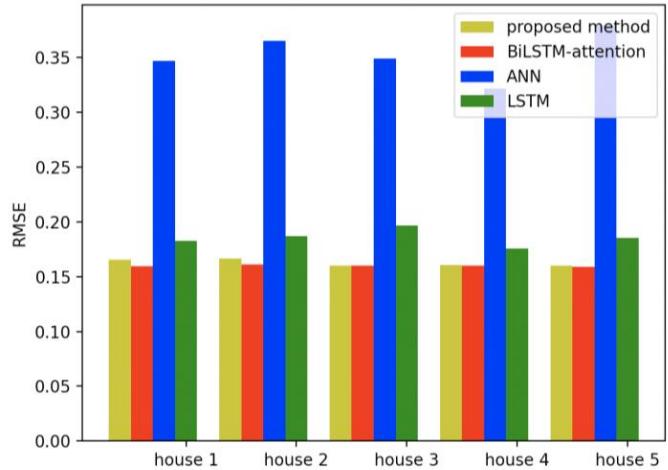

Fig. 21. Comparison of 5 households using RMSE using different methods

### 4.4 Robustness analysis of federated Settings

To evaluate the effect of primary parameters on the robustness of the proposed FedDL prediction model, we design a different experiment with varied local Iteration period E and the client score C that performs calculations during training. Here, the number of local iteration cycles for each round of clients is set to 5, 50, and 80; the scores for the number of clients are set to 50% and 100%. The results are displayed in Fig. 22.

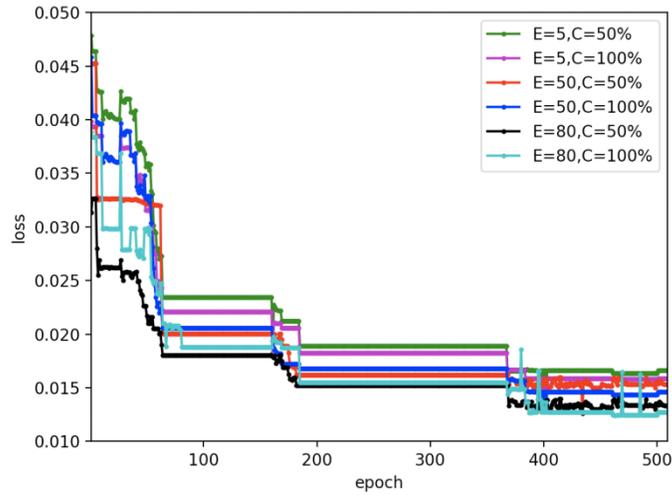

Fig. 22. Loss results under different federal settings

As can be seen from Fig. 22, different federal settings have different effects on FedDL performance. When the local client epochs are fixed, the performance of the model would not change with different local client epochs, which proved that the prediction model has good robustness to different local client epochs. And when the client ratio is fixed, the loss function value of the prediction model reduces slightly with the increase of iterations.

## 5. Conclusions

This paper proposes a household load prediction method based on federated deep learning and non-intrusive load monitoring. Based on the simulation results, the following conclusions can be drawn:

(1) By combining federal learning with BiLSTM-Attention, this work develops a household load prediction approach based on non-intrusive load monitoring via federated deep learning. In this prediction method, under the precondition of guaranteeing the accuracy of the prediction, the data of household users are not leaked. It can reduce the data security risk and use the electrical power of other households to meet the needs for user privacy protection.

(2) The NILM based on CNN-LSTM hybrid model is used to decompose the individual load of household appliances from the overall load demand, and the power of each appliance is predicted by the decomposed power of individual appliances, which avoids the strong time dependence of individual equipment. The case results demonstrate that this NILM-based integrated prediction methodology provides a better prediction effect than the traditional methodology that directly predicts the aggregated signal as a whole.

(3) This paper uses the data decomposed by the CNN-LSTM hybrid deep learning model to test the proposed federated deep learning prediction model. The load forecasting result is very close to the real loads. And through comparison with other models, the superiority of the proposed method is verified. In addition, experiments in various federated learning environments are implemented to validate the robustness of this methodology.

In this paper, external influence factors, such as weather factors and humidity factors, are not taken into account in the prediction process, but these factors need to be considered in more realistic scenarios. It is interesting to extend the proposed NILM algorithm for supporting the energy management of residential integrated energy systems [50, 51]. Besides, using automated reinforcement learning for determining the optimal hyper parameters of the used deep learning model [52] is also deserved to be investigated. Another interesting topic is to investigate the NILM issues under cyber-attacks [53-55].

# References

[1]  López G, Custodio V, Moreno J I, et al. Modeling smart grid neighborhoods with the ENERsip ontology[J]. Computers In Industry, 2015, 70: 168-182.

[2]  Wu Z, Zhou S, Li J, et al. Real-time scheduling of residential appliances via conditional risk-at-value[J]. IEEE Transactions on Smart Grid, 2014, 5(3): 1282-1291.

[3]  Li Y, Li K, Yang Z, et al. Stochastic optimal scheduling of demand response-enabled microgrids with renewable generations: An analytical-heuristic approach[J]. Journal of Cleaner Production, 2022, 330: 129840.

[4]  Li Y, Wang B, Yang Z, et al. Optimal scheduling of integrated demand response-enabled community-integrated energy systems in uncertain environments[J]. IEEE Transactions on Industry Applications, 2022, 58(2): 2640-2651.

[5]  Wang J, Jia R, Zhao W, et al. Application of the largest Lyapunov exponent and non-linear fractal extrapolation algorithm to short-term load forecasting[J]. Chaos, Solitons & Fractals, 2012, 45(9-10): 1277-1287.

[6]  Nazih A S, Fawwaz E, Osama M A. Medium-term electric load forecasting using multivariable linear and non-linear regression[J]. Smart Grid and Renewable Energy, 2011, 2011.

[7]  Sadaei H J, e Silva P C L, Guimaraes F G, et al. Short-term load forecasting by using a combined method of convolutional neural networks and fuzzy time series[J]. Energy, 2019, 175: 365-377.

[8]  Aly H H H. A proposed intelligent short-term load forecasting hybrid models of ANN, WNN and KF based on clustering techniques for smart grid[J]. Electric Power Systems Research, 2020, 182: 106191.

[9]  Gasparin A, Lukovic S, Alippi C. Deep learning for time series forecasting: The electric load case[J]. CAAI Transactions on Intelligence Technology, 2022, 7(1): 1-25.

[10]  Wang Y, Xia Q, Kang C. Secondary forecasting based on deviation analysis for short-term load forecasting[J]. IEEE Transactions on Power Systems, 2010, 26(2): 500-507.

[11]  Zeng J, Qiao W. Short-term solar power prediction using a support vector machine[J]. Renewable Energy, 2013, 52: 118-127.

[12]  Yildiz B, Bilbao J I, Sproul A B. A review and analysis of regression and machine learning models on commercial building electricity load forecasting[J]. Renewable and Sustainable Energy Reviews, 2017, 73: 1104-1122.

[13]  Cui M, Wang J, Yue M. Machine learning-based anomaly detection for load forecasting under cyberattacks[J]. IEEE Transactions on Smart Grid, 2019, 10(5): 5724-5734.

[14]  Kumar S, Mishra S, Gupta S. Short term load forecasting using ANN and multiple linear regression[C]//2016 second international conference on computational intelligence & communication technology (cict). IEEE, 2016: 184-186.

[15]  Shi Z, Li Y, Yu T. Short-term load forecasting based on LS-SVM optimized by bacterial colony chemotaxis algorithm [C]. International Conference on Information and Multimedia Technology, Jeju Island, South Korea, 2009.

[16]  Yang Li, Xueping Gu. Application of online SVR in very short-term load forecasting. International Review of Electrical Engineering (IREE), 2013, 8(1): 277-282.

[17]  Liu Q, Shen Y, Wu L, et al. A hybrid FCW-EMD and KF-BA-SVM based model for short-term load forecasting[J]. CSEE Journal of Power and Energy Systems, 2018, 4(2): 226-237.

[18]  Ziel F, Liu B. Lasso estimation for GEFCom2014 probabilistic electric load forecasting[J]. International Journal of Forecasting, 2016, 32(3): 1029-1037.

[19]  Chen B J, Chang M W. Load forecasting using support vector machines: A study on EUNITE competition 2001[J]. IEEE Transactions on Power Systems, 2004, 19(4): 1821-1830.

[20]  Stephen B, Tang X, Harvey P R, et al. Incorporating practice theory in sub-profile models for short term aggregated residential load forecasting[J]. IEEE Transactions on Smart Grid, 2015, 8(4): 1591-1598.

[21] Guan C, Luh P B, Michel L D, et al. Hybrid Kalman filters for very short-term load forecasting and prediction interval estimation[J]. IEEE Transactions on Power Systems, 2013, 28(4): 3806-3817.

[22] Sharma S, Majumdar A, Elvira V, et al. Blind Kalman filtering for short-term load forecasting[J]. IEEE Transactions on Power Systems, 2020, 35(6): 4916-4919.

[23] Zheng Z, Chen H, Luo X. A Kalman filter-based bottom-up approach for household short-term load forecast[J]. Applied Energy, 2019, 250: 882-894.

[24] Lee C M, Ko C N. Short-term load forecasting using lifting scheme and ARIMA models[J]. Expert Systems with Applications, 2011, 38(5): 5902-5911.

[25] Kuster C, Rezgui Y, Mourshed M. Electrical load forecasting models: A critical systematic review[J]. Sustainable cities and society, 2017, 35: 257-270.

[26] Wang Y, Chen Q, Sun M, et al. An ensemble forecasting method for the aggregated load with subprofiles[J]. IEEE Transactions on Smart Grid, 2018, 9(4): 3906-3908.

[27] Ponoćko J, Milanović J V. Forecasting demand flexibility of aggregated residential load using smart meter data[J]. IEEE Transactions on Power Systems, 2018, 33(5): 5446-5455.

[28] Li L, Meinrenken C J, Modi V, et al. Short-term apartment-level load forecasting using a modified neural network with selected auto-regressive features[J]. Applied Energy, 2021, 287: 116509.

[29] Fang Z, Zhao D, Chen C, et al. Nonintrusive appliance identification with appliance-specific networks. IEEE Transactions on Industry Applications[J]. 2020, 56(4):3443-3452.

[30] Dinesh C, Makonin S, Bajić I V. Residential power forecasting based on affinity aggregation spectral clustering[J]. IEEE Access, 2020, 8: 99431-99444.

[31] Wang Y, Bennani I L, Liu X, et al. Electricity consumer characteristics identification: A federated learning approach[J]. IEEE Transactions on Smart Grid, 2021.

[32] Li L, Fan Y, Tse M, et al. A review of applications in federated learning[J]. Computers & Industrial Engineering, 2020: 106854.

[33] Li Y, Li J, Wang Y. Privacy-preserving spatiotemporal scenario generation of renewable energies: A federated deep generative learning approach[J]. IEEE Transactions on Industrial Informatics, 2022, 18 (4): 2310-2320.

[34] Yang Z, Chen M, Saad W, et al. Energy efficient federated learning over wireless communication networks[J]. IEEE Transactions on Wireless Communications, 2020, 20(3): 1935-1949.

[35] Briggs C, Fan Z, Andras P. Federated learning for short-term residential energy demand forecasting[J]. ArXiv Preprint ArXiv:2105.13325, 2021.

[36] Shi Z, Yu T, Zhao Q, et al. Comparison of algorithms for an electronic nose in identifying liquors. Journal of Bionic Engineering[J]. 2008, 5(3), 253-257.

[37] Lun X, Jia S, Hou Y, et al. GCNs-net: a graph convolutional neural network approach for decoding time-resolved EEG motor imagery signals[J]. 2020, arXiv preprint arXiv:2006.08924.

[38] Lawrence S, Giles C L, Tsoi A C, et al. Face recognition: A convolutional neural-network approach[J]. IEEE Transactions on Neural Networks, 1997, 8(1): 98-113.

[39] Lindemann B, Maschler B, Sahlab N, et al. A survey on anomaly detection for technical systems using LSTM networks[J]. Computers In Industry, 2021, 131: 103498.

[40] Zhang M, Li J, Li Y, et al. Deep learning for short-term voltage stability assessment of power systems [J]. IEEE Access, 2021, 9, 29711-29718.

[41] Liu G, Guo J. Bidirectional LSTM with attention mechanism and convolutional layer for text classification[J]. Neurocomputing, 2019, 337: 325-338.

[42] Liu S, Lee K, Lee I. Document-level multi-topic sentiment classification of email data with bilstm and data augmentation[J]. Knowledge-Based

Systems, 2020, 197: 105918.

[43] Bin Y, Yang Y, Shen F, et al. Describing video with attention-based bidirectional LSTM[J]. IEEE Transactions on Cybernetics, 2018, 49(7): 2631-2641.

[44] Ji Z, Xiong K, Pang Y, et al. Video summarization with attention-based encoder–decoder networks[J]. IEEE Transactions on Circuits and Systems for Video Technology, 2019, 30(6): 1709-1717.

[45] Li Y, Zhang M, Chen C. A deep-learning intelligent system incorporating data augmentation for short-term voltage stability assessment of power systems[J]. Applied Energy, 2022, 308, 118347.

[46] Huong T T, Bac T P, Long D M, et al. Detecting cyberattacks using anomaly detection in industrial control systems: A Federated Learning approach[J]. Computers In Industry, 2021, 132: 103509.

[47] Lu Y, Huang X, Dai Y, et al. Blockchain and federated learning for privacy-preserved data sharing in industrial IoT[J]. IEEE Transactions on Industrial Informatics, 2019, 16(6): 4177-4186.

[48] Zhou X, Feng J, Li Y. Non-intrusive load decomposition based on CNN–LSTM hybrid deep learning model[J]. Energy Reports, 2021, 7: 5762-5771.

[49] Kelly J, Knottenbelt W. The UK-DALE dataset, domestic appliance-level electricity demand and whole-house demand from five UK homes[J]. Scientific Data, 2015, 2(1): 1-14.

[50] Sun J, Deng J, Li Y. Indicator & crowding distance-based evolutionary algorithm for combined heat and power economic emission dispatch[J]. Applied Soft Computing, 2020, 90: 106158.

[51] Li Y, Wang B, Yang Z, et al. Hierarchical stochastic scheduling of multi-community integrated energy systems in uncertain environments via Stackelberg game[J]. Applied Energy, 2022, 308: 118392.

[52] Li Y, Wang R, Yang, Z. Optimal scheduling of isolated microgrids using automated reinforcement learning-based multi-period forecasting[J]. IEEE Transactions on Sustainable Energy, 2022, 13(1), 159-169.

[53] Wang J, Srikantha P. Stealthy black-box Attacks on deep learning non-intrusive load monitoring models. IEEE Transactions on Smart Grid, 2021, 12(4), 3479-3492.

[54] Li Y, Li Z, Chen, L. Dynamic state estimation of generators under cyber attacks[J]. IEEE Access, 2019, 7, 125253-125267.

[55] Iliaee N, Liu S, Shi W. Non-Intrusive load monitoring based demand prediction for smart meter attack detection[C]. In 2021 International Conference on Control, Automation and Information Sciences (ICCAIS), pp. 370-374, IEEE, 2021.